\documentclass[sigconf]{acmart-arxiv}

\usepackage{multirow}
\usepackage{enumitem}
\usepackage{placeins}
\usepackage{tikz}
\usetikzlibrary{arrows.meta,positioning,shapes.geometric}

\title[Operational Runbooks as Living Documents]{Operational Runbooks as Living Documents:
 A Longitudinal Study of Document Evolution under Assisted Post-Incident Revision}

\author{Rocker D'Antonio}
\orcid{0009-0001-8444-1772}
\affiliation{
  \institution{Amazon Web Services}
  \city{Santa Cruz}
  \state{California}
  \country{USA}
}
\affiliation{
  \institution{Mississippi State University}
  \city{Mississippi State}
  \state{Mississippi}
  \country{USA}
}

\author{Harry Xie}
\orcid{0009-0008-5946-9328}
\affiliation{
  \institution{Amazon Web Services}
  \city{Seattle}
  \state{Washington}
  \country{USA}
}

\author{Xuan Guo}
\orcid{0009-0006-8864-4709}
\affiliation{
  \institution{Amazon Web Services}
  \city{Seattle}
  \state{Washington}
  \country{USA}
}

\copyrightyear{2026}
\acmYear{2026}
\setcopyright{cc}
\setcctype{by}
\acmConference[DocEng '26]{Proceedings of the 2026 ACM Symposium on Document Engineering}{August 25--28, 2026}{Fribourg, Switzerland}
\acmBooktitle{Proceedings of the 2026 ACM Symposium on Document Engineering (DocEng '26), August 25--28, 2026, Fribourg, Switzerland}
\acmDOI{10.1145/3820755.3821485}
\acmISBN{979-8-4007-2786-3/2026/08}

\begin{document}
\begin{abstract}
Operational runbooks increasingly function as living documents within operational workflows: they are maintained by people, used in incident support, and continuously revised as organizational knowledge changes. Yet little is known about how such document collections evolve over time in production settings, or which interpretable signals are useful for monitoring document change. We analyze 17 weeks of version-controlled runbook snapshots produced during post-incident revision with machine-assisted review and human oversight. We quantify document evolution using lightweight linguistic and structural signals grounded in prior work on procedural language, technical communication, and document organization. Correlation analysis shows mixed patterns of association, including both positive and negative relationships, with effect sizes ranging from negligible to moderate and varying across corpora. Temporal summaries further distinguish comparatively stable signals from revision-sensitive ones, supporting monitoring of document maintenance workflows rather than one-shot quality judgments. An illustrative analysis on public GitHub product documentation shows that signal behavior differs across corpora, reinforcing the need for context-aware interpretation. These results identify which metric classes provide stable anchors and which provide edit-sensitive indicators for managing operational documentation as an engineered document resource.
\end{abstract}



\maketitle

\section{Introduction}

Operational documentation occupies a distinct position among software engineering artifacts. Runbooks are procedural documents that provide instructions for operational tasks such as alarm triage, incident response, and maintenance activities in deployed systems. Unlike descriptive documentation intended primarily for learning or reference, runbooks and related support materials guide actions whose timeliness and correctness can directly affect service reliability. Authors must therefore balance thoroughness against navigability, preserving critical detail without impeding rapid information access.

While care and upkeep practices for these materials vary between institutions, the current standard is trending towards managed document collections embedded within version-controlled documentation workflows, a practice known in industry vernacular as 'doc-as-code'. Such repositories evolve through recurring cycles of authoring, review, approval, and reuse. Documentation revision is often uneven and event-driven, producing variability that complicates aggregate analysis. We therefore frame revision monitoring as a document engineering problem: maintainers need lightweight methods to track how documents change across revisions. The practical question is not which signals are universally stable, but which can be meaningfully interpreted under these conditions. 

Answering this question requires understanding how commonly used document analysis metrics behave as documentation evolves. Prior work on technical documentation often emphasizes static quality assessment or downstream task outcomes, offering less guidance on the behavior of metrics across successive revisions. As a result, maintainers have limited support for distinguishing between routine variation, meaningful document change, and metric instability in evolving repositories. 

We look at documentation revision after incidents to examine how different metric classes behave under routine maintenance pressure. Some metrics remain comparatively stable and support baseline monitoring. Others are responsive to localized reformulation and help identify where revision activity is concentrated. A third group is best treated as context-dependent because similar metric changes may require different interpretations depending on repository context.

Our study analyzes two documentation corpora. The first consists of operational runbooks produced during a 17-week deployment previously described in prior industry work \citep{DantonioXie2025}, where documentation was revised through routine post-incident refinement supported by automated gap analysis tools and human review. In this setting, automated assistance was used to bring to light potential gaps, while authors remained responsible for all revisions. We re-examine the version-controlled artifacts from that setting as a documentation repository with repeated snapshots, revision history, and established maintenance practices. The second corpus consists of yearly snapshots from a public technical documentation repository and provides a comparison point for how the same metrics behave under a different collection scale and revision regime.

Since this work focuses on documentation change rather than downstream task performance, increases or decreases in features such as hedging, modal language, and constraint expressions are not interpreted as improvements or degradations in document quality. Instead, these features are examined as characteristics of repository evolution. This scope is particularly important when documentation serves both human readers and automated systems, which may rely on different mechanisms for interpreting textual characteristics.

\emph{Contributions.}
This paper makes three contributions. First, it defines a lightweight measurement approach for monitoring documentation revision in version-controlled operational documentation using transparent linguistic and structural metrics. Second, it introduces a stability-oriented metric categorization that distinguishes stable baseline measures from revision-sensitive indicators under routine post-incident refinement in a deployed operational setting, while also identifying context-dependent metrics whose interpretation depends on repository conditions. Third, through comparison with a public technical documentation corpus, it shows that metric behavior differs across the two corpora studied. The paper focuses on measurement and interpretation of documentation revision rather than validation against downstream operational outcomes.

\section{Related Work}
\label{sec:Related Work}

Research on technical documentation spans natural language processing, software engineering, and technical communication, but often treats documentation as a static artifact or evaluates interventions indirectly through downstream outcomes, leaving open how documentation language itself evolves over time.

\textit{Evaluation in deployed operational settings} highlights challenges associated with uncontrolled environments, where event-driven behavior introduces variability and complicates causal attribution. Prior work therefore emphasizes descriptive and distributional analyses, within-subject comparisons, and transparent evaluation signals that remain stable under distribution shift \citep{chen2023menli}. In industry contexts, rubric-based frameworks incorporating expert judgment have been proposed for domain-specific documentation quality, particularly in clinical text \citep{shah2025tn}. This study applies metrics to documentation artifacts in an operational setting, where the scale of ongoing updates makes continuous human labeling impractical.

\textit{Documentation quality for retrieval-augmented and related documentation dependent systems} has also been examined in the context of systems that rely on external documentation for retrieval and grounding. Recent work shows that properties such as coverage, organization, and specificity influence retrieval behavior and downstream task outcomes, and proposes evaluation frameworks based on task-level performance or model-based judgments \citep{wang-etal-2024-searching, saadfalcon2024aresautomatedevaluationframework, wijaya2025readmellmframeworkhelpllms}. A recent analysis of language-model-generated documentation defines quality along dimensions such as accuracy, completeness, relevance, and understandability, assessed through expert review or learned evaluators, and serves as an example of evaluation framed around generated output quality rather than properties of the documentation itself \citep{dvivedi2024comparativeanalysislargelanguage}. In our work, we focus on the documentation language itself. We use lightweight linguistic and structural signals to examine how documentation changes over time under post-incident refinement in an operational setting. The analysis here focuses on identifying metrics whose behavior can be interpreted as documentation is incrementally revised.

\textit{Documentation growth and structural organization} have been examined using quantitative properties such as document length, section count, and content growth in document analysis and information retrieval \citep{SparckJones1972,Aizawa2003}. In technical communication, structural organization through headings, hierarchical sections, and consistent formatting has been shown to support navigability and comprehension. In NLP, document structure is leveraged for tasks such as summarization and retrieval, motivating the use of header density and section-level statistics as interpretable structural signals.

\textit{Procedural and instructional language} has been studied in linguistics and in computational NLP. Researchers have treated procedural text as a distinct understanding task, modeling how entities and states evolve across step sequences \citep{tang2020understanding}, and examining causal reasoning over entities and events in procedural narratives \citep{zhang2023causal}. Earlier research on procedural knowledge extraction demonstrates how surface linguistic cues can be used to identify and structure procedural content automatically \citep{zhang2012automatically}. Across these studies, features such as imperative constructions, step enumeration, and explicit constraints are recognized as indicators of procedural intent and actionability. The empirical patterns observed in our analysis are consistent with these frameworks, in that changes in imperative framing and instructional structure provide interpretable signals of how procedural content is revised over time in operational documentation.

\textit{Concreteness, parameterization, and referential grounding} are commonly operationalized through surface-level indicators such as numeric expressions, parameters, identifiers, command snippets, named entities, and pronoun usage. These features are widely used in software documentation analysis, information retrieval, readability assessment, and text complexity research as transparent proxies for operational detail and task specificity \citep{stajner-hulpus-2018-automatic, davoodi2017discourse, imperial2021application, Nadeau2007, sahay2025aspects}.

\textit{Vagueness and hedging} have been studied in linguistic pragmatics and applied NLP, with hedge terms and epistemic modals introducing uncertainty and variability in interpretation, often reducing directive clarity \citep{goodman2014probabilistic, de-marneffe-etal-2012-happen}. Lexically grounded and interpretable methods such as SpeciTeller are commonly used in exploratory analyses that prioritize robustness and transparency under operational variability \citep{li2015speciteller}.

\section{Study Context and Data}
\label{sec:context_data}

This study analyzes documentation produced during a previously reported deployment for operational incident response \citep{DantonioXie2025}. In that setting, documentation refinement occurred through a review process in which automated gap analysis surfaced candidate updates for human review. The present work introduces no new data sources and does not modify the original deployment; instead, it re-examines the documentation artifacts from that deployment through longitudinal language analysis.

\emph{Operational context.}
The deployment supported on-call engineers by retrieving documentation and presenting synthesized guidance during incident handling. Following incident resolution, the guidance initially surfaced was compared with documented resolution steps to identify discrepancies indicative of missing or underspecified documentation. These discrepancies were then surfaced as refinement suggestions for human review and incorporation through established documentation-as-code workflows. Suggestions ranged from concrete procedural additions to higher-level identification of documentation gaps. All documentation updates remained human-authored and subject to standard review and approval processes.

\begin{figure}[t]
\centering
\small
\begin{tikzpicture}[
  node distance=0.26cm,
  box/.style={
    rectangle,
    rounded corners,
    draw=black,
    align=center,
    text width=0.50\columnwidth,
    minimum height=0.48cm,
    inner sep=1.6pt
  },
  data/.style={
    trapezium,
    trapezium left angle=75,
    trapezium right angle=105,
    draw=black,
    align=center,
    text width=0.50\columnwidth,
    minimum height=0.48cm,
    inner sep=1.6pt
  },
  arrow/.style={-{Latex[length=1.6mm]}, semithick}
]
\node[data] (incident) {Incident context\\and telemetry};
\node[box, below=of incident] (agent) {Ops agent retrieves\\runbooks, surfaces guidance};
\node[box, below=of agent] (operator) {Engineer uses guidance\\during incident response};
\node[box, below=of operator] (compare) {Compare surfaced guidance\\with documented resolution};
\node[box, below=of compare] (suggest) {Generate gap analysis\\and revision suggestions};
\node[box, below=of suggest] (review) {Human review and\\docs-as-code approval};
\node[data, below=of review] (runbooks) {Version-controlled\\runbook repository};

\draw[arrow] (incident) -- (agent);
\draw[arrow] (agent) -- (operator);
\draw[arrow] (operator) -- (compare);
\draw[arrow] (compare) -- (suggest);
\draw[arrow] (suggest) -- (review);
\draw[arrow] (review) -- (runbooks);
\draw[arrow] (runbooks.east) .. controls +(0.28,0) and +(0.28,0) .. (agent.east);
\end{tikzpicture}
\caption{Operational workflow for runbook use and revision, in which agent-surfaced guidance is compared with incident resolution to identify gaps that are reviewed and incorporated into the version-controlled repository.}
\Description{A vertical workflow starts with incident context and telemetry, proceeds through agent retrieval of runbooks, engineer use during incident response, comparison with documented resolution, generation of gap analysis and revision suggestions, human review and docs-as-code approval, and ends at a version-controlled runbook repository that feeds back into agent retrieval.}
\label{fig:workflow}
\end{figure}
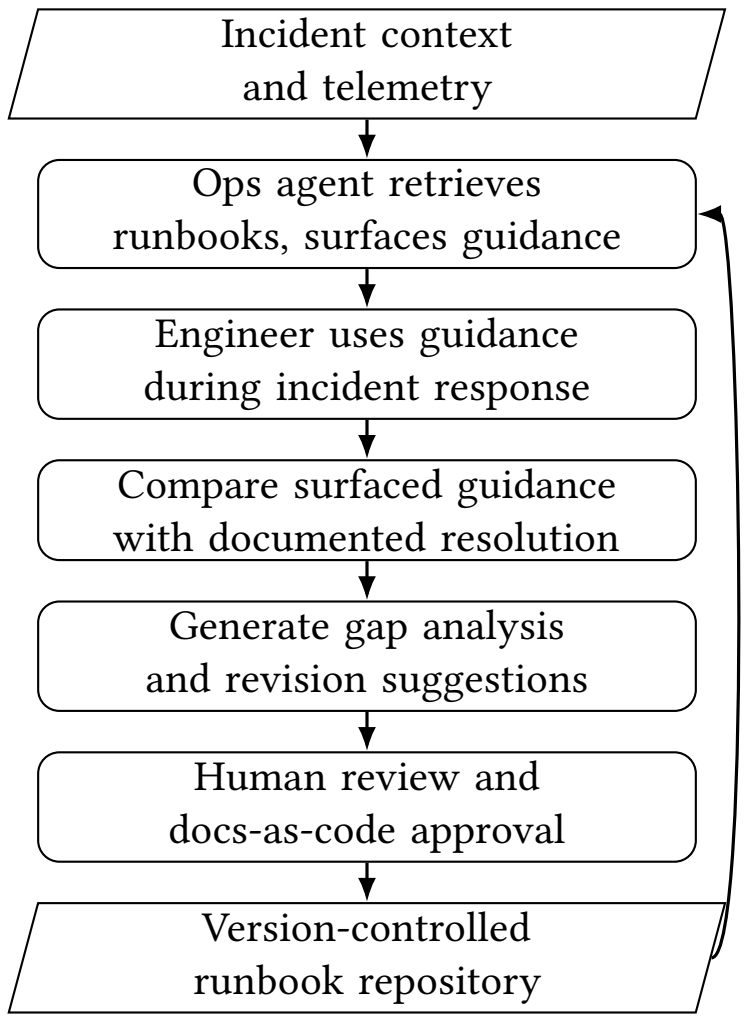

\emph{Documentation corpus and temporal structure.}
Operational documentation was maintained as a collection of Markdown files in a shared version-controlled repository, including architectural documents, procedural runbooks and SOPs, incident triage guides, and related operational references. Documentation changes were recorded through commits and segmented into weekly snapshots, each representing the state of the documentation at the end of a given week-long on-call operator shift. This snapshot-based representation supports longitudinal comparison while reducing sensitivity to short-term variation in commit activity. The dataset spans a fixed 17-week period corresponding to the original deployment. Because many observations correspond to repeated measurements of the same documents across snapshots (i.e., pairs of documents and snapshots), we treat the dataset as longitudinal rather than as a collection of independent samples.

In addition to the operational corpus, we analyze a public technical documentation repository using yearly snapshots. The public corpus is included as a reference setting for applying the same measurements outside the operational deployment. It is not used as a controlled comparison and differs from the operational corpus in scale, revision cadence, and document role.

\section{Methodology}
\label{Methodology}

Our goal is to characterize how operational documentation changes over time when updates are introduced during assisted post-incident revision. Using version-controlled snapshots, we analyze a set of lightweight linguistic and structural metrics as monitoring signals for how revisions accumulate within a maintained documentation repository.

\subsection{Unit of Analysis}

The primary unit of analysis is a weekly documentation snapshot reconstructed from a version-controlled repository, representing the state of all documents at the end of each week. For each snapshot, metrics are aggregated to produce corpus-level distributions used in the analysis. In addition, changes are tracked within individual documents across weeks to support paired comparisons, reducing confounding effects due to document type or baseline verbosity.

\subsection{Metric Selection Rationale}

Metrics are selected to satisfy three practical constraints: they must be supported by prior linguistic or technical communication research, computable using lightweight and transparent procedures, and suitable for longitudinal and distributional analysis in production data without requiring controlled experimental conditions. The resulting metric set spans a range of linguistic and structural properties relevant to technical documentation and is analyzed in aggregate and over time.

\subsection{Linguistic and Structural Metrics}

\textit{Content volume and baseline activity.} We compute basic volume-related measures, including word count and sentence count per document, to contextualize documentation growth over time. These metrics are not treated as indicators of documentation quality but are used to normalize and interpret subsequent measures.

\textit{Structural organization.} Structural properties are measured using features derived from document formatting and sectioning, including header counts, header depth distributions, and section-level statistics. These measures capture changes in navigability and organizational regularity without semantic interpretation.

\textit{Procedural explicitness.} Procedural explicitness is operationalized using surface indicators of instructional language, including the density of enumerated steps and the rate of imperative sentence constructions. These metrics provide interpretable signals of procedural intent while remaining robust to stylistic variation.

\textit{Concreteness and parameterization.} We measure the presence of explicit operational detail using numeric expression density and the frequency of code blocks and inline command tokens, which commonly convey executable actions in technical documentation.

\textit{Referential grounding.} Referential grounding is approximated using named entity density, entity diversity, and pronoun usage rates, reflecting the extent to which documentation relies on concrete references rather than underspecified terms.

\textit{Vagueness and constraint language.} Uncertainty and obligation are measured using curated lexical markers of vagueness (e.g. "may", "likely" ) and constraint (e.g. "required", "never") then computed as density measures to enable cross-document comparison. The lexical proxies used for vagueness and constraint may exhibit partial overlap and are interpreted as approximate signals rather than strictly independent measures. Full marker lists are provided in Appendix~\ref{app:metrics}.

All metrics were computed from Markdown source using deterministic rules and standard NLP preprocessing (tokenization, sentence segmentation, POS tagging, and NER) implemented with spaCy; exact metric definitions and marker sets are provided in Appendix~\ref{app:metrics}.

\subsection{Statistical summaries}

To characterize relationships among key linguistic signals, we compute Spearman rank correlations between imperative verb ratio, constraint density, vagueness density, and entity density within each corpus. Correlations are computed across document-level observations within snapshots and reflect cross-sectional associations rather than temporal co-movement. As a result, signals may exhibit positive correlation at the snapshot level while moving in opposite directions over time. These views capture different aspects of documentation behavior and should not be interpreted as contradictory. We report both $\rho$ and the associated $p$-value, but effect size is the primary focus. Given repeated observations across snapshots, $p$-values are reported for completeness but are not interpreted as evidence of statistical significance. To summarize temporal behavior in the longitudinal HIL corpus, we additionally report the coefficient of variation (CV) over weekly aggregate means, using lower CV values to identify stable anchors, higher values to identify revision-sensitive indicators, and intermediate values to motivate context-dependent interpretation.

\section{Results}
\label{sec:results}

\subsection{Corpus-Level Descriptive Statistics}
\label{sec:results:descriptive}

Over the 17-week revision timeline, the corpus grew from 57 to 78 documents through 28 reviewed documentation changes, representing a 36.8\% increase in document count and a 36.2\% increase in total word volume (51,920 to 70,703 words). Despite this growth, document length remained stable: mean words per document declined slightly ($-0.5$\%), while the median increased modestly (+7.7\%). Mean sentence count per document decreased ($-2.9$\%), and average sentence length remained approximately constant at 26.4 tokens.

\IfFileExists{tables/tab_weekly_mean_range.tex}{\begin{table}[t]
\caption{Range of weekly aggregate means across the 17-week study timeline (computed from weekly snapshot summaries).}
\Description{Table listing six longitudinal metrics with minimum, median, and maximum values across the 17-week study. Rows include valid documents, entity density, imperative verb ratio, vagueness density, constraint density, and code fence count.}
\label{tab:weekly_mean_range}
\centering
\small
\begin{tabular}{lrrr}
\toprule
Metric & Min & Median & Max \\
\midrule
Valid Documents & 57 & 73 & 78 \\
Entity Density (Mean) & 0.06 & 0.06 & 0.06 \\
Imperative Verb Ratio (Mean) & 0.06 & 0.07 & 0.08 \\
Vagueness Density (Mean) & 7.94 & 8.67 & 8.91 \\
Constraint Density (Mean) & 6.13 & 7.25 & 7.57 \\
Code Fence Count (Mean) & 1.06 & 1.23 & 1.29 \\
\bottomrule
\end{tabular}
\end{table}
}{\IfFileExists{tables/tab_weekly_mean_range}{}{}}

\IfFileExists{tables/tab_weekly_mean_range.tex}{Table~\ref{tab:weekly_mean_range} reports the range of weekly aggregate means for key metrics.}{\IfFileExists{tables/tab_weekly_mean_range}{Table~\ref{tab:weekly_mean_range} reports the range of weekly aggregate means for key metrics.}{The weekly aggregate means for key metrics show only limited variability across the revision window.}} Several metrics exhibited limited variability, while others showed moderate to higher variation: entity density remained tightly bounded around 0.06, imperative verb ratio ranged from 0.06 to 0.08, and vagueness and constraint densities showed bounded variation. Code fence count per document ranged from 1.06 to 1.29. Full endpoint values and per-metric summaries are provided in Appendix~\ref{app:descriptive}.

Overall, the corpus exhibited steady growth in document count while maintaining stable document length and bounded linguistic variation.

\subsection{Week-over-Week Trends}
\label{sec:results:trends}

\begin{table}[t]
\caption{HIL focal-signal movement from W1 to W17.}
\Description{Table comparing selected HIL metrics at Week 1 and Week 17 with net percentage change. Entity density is nearly stable, imperative verb ratio decreases, and vagueness and constraint density increase.}
\label{tab:hil_signal_movement}
\centering
\small
\setlength{\tabcolsep}{5pt}
\begin{tabular}{lccc}
\toprule
\textbf{Metric} & \textbf{W1} & \textbf{W17} & \textbf{$\Delta$\%} \\
\midrule
Entity density & 0.063 & 0.061 & -3.5\% \\
Imperative verb ratio & 0.084 & 0.065 & -23.2\% \\
Vagueness density & 8.11 & 8.89 & +9.7\% \\
Constraint density & 6.13 & 7.25 & +18.2\% \\
\bottomrule
\end{tabular}
\end{table}

Across the revision period, metrics changed gradually but did not follow a uniform direction. Entity density showed limited variation relative to vagueness, varying by at most $-3.5$\% from baseline, while the imperative verb ratio declined steadily, reaching $-23.2$\% by Weeks~13 and~17. In contrast, vagueness density increased by 9.7\% and constraint density by 18.2\%.

Procedural structure indicators showed fluctuating behavior without sustained directional trends. Code fence count declined early before partially recovering and stabilizing near baseline, while list items per document exhibited small fluctuations followed by a slight net decrease.

Table~\ref{tab:hil_signal_movement} makes the core HIL signal movement explicit at a glance: referential grounding changed little, imperative framing declined, and both vagueness and constraint language increased. Across metrics, trajectories evolved independently: entity density remained comparatively bounded while constraint density increased, imperative framing declined as vagueness increased, and structural indicators varied without consistent correlation to volume measures. These endpoint shifts indicate direction of change, but not its smoothness across weekly snapshots. Detailed weekly percentage changes are reported in Appendix~\ref{app:hil_trends}.

\subsection{Correlation structure across linguistic signals}

\begin{table}[t]
\caption{Spearman correlations among key linguistic signals. Effect labels are included to discourage overinterpretation of small associations.}
\Description{Table of Spearman correlations for signal pairs in the HIL and GitHub corpora, with rho, p-value, and qualitative magnitude. It shows mostly weak associations with a few moderate relationships.}
\label{tab:signal_correlations}
\centering
\footnotesize
\setlength{\tabcolsep}{4pt}
\begin{tabular}{llccc}
\toprule
\textbf{Corpus} & \textbf{Signal pair} & $\boldsymbol{\rho}$ & \textbf{$p$-value} & \textbf{Magnitude} \\
\midrule
HIL & imperative $\leftrightarrow$ constraint & 0.16 & $<0.001$ & weak positive \\
HIL & imperative $\leftrightarrow$ vagueness & 0.13 & $<0.001$ & weak positive \\
HIL & imperative $\leftrightarrow$ entity & 0.31 & $<0.001$ & moderate positive \\
HIL & constraint $\leftrightarrow$ vagueness & 0.33 & $<0.001$ & moderate positive \\
HIL & constraint $\leftrightarrow$ entity & -0.17 & $<0.001$ & weak negative \\
HIL & vagueness $\leftrightarrow$ entity & -0.38 & $<0.001$ & moderate negative \\
\midrule
GitHub & imperative $\leftrightarrow$ constraint & 0.21 & $<0.001$ & weak positive \\
GitHub & imperative $\leftrightarrow$ vagueness & 0.04 & $<0.001$ & negligible \\
GitHub & imperative $\leftrightarrow$ entity & 0.15 & $<0.001$ & weak positive \\
GitHub & constraint $\leftrightarrow$ vagueness & 0.13 & $<0.001$ & weak positive \\
GitHub & constraint $\leftrightarrow$ entity & 0.01 & 0.392 & negligible \\
GitHub & vagueness $\leftrightarrow$ entity & -0.26 & $<0.001$ & weak negative \\
\bottomrule
\end{tabular}
\end{table}

Table~\ref{tab:signal_correlations} shows mixed patterns of association, including both positive and negative relationships. In the HIL corpus, the strongest relationships were a moderate positive association between constraint and vagueness density ($\rho=0.33$), a moderate negative association between vagueness and entity density ($\rho=-0.38$), and a moderate positive association between imperative ratio and entity density ($\rho=0.31$). By contrast, imperative--constraint ($\rho=0.16$) and imperative--vagueness ($\rho=0.13$) were weak positive associations, while constraint--entity remained weak overall. Correlation summaries should be read together with the aggregate trend analysis above: cross-sectional association within snapshots and aggregate movement over time capture different aspects of revision behavior.

Because documents are observed repeatedly across snapshots, these row-level correlations likely overstate the effective sample size in the HIL corpus. We therefore interpret the correlations conservatively, prioritize magnitude over $p$-values, and emphasize recurring patterns rather than weak effects. The GitHub corpus shows a different configuration. Imperative ratio and constraint density remained weakly positively associated ($\rho=0.21$), and vagueness and entity density showed a weak negative association ($\rho=-0.26$). However, imperative--vagueness was negligible ($\rho=0.04$) and constraint--entity was effectively null ($\rho=0.01$). These cross-corpus differences indicate that relationships among directive, pragmatic, and referential signals differ across the two corpora studied.

\begin{table}[t]
\caption{Cross-corpus comparison of focal signal movement over the observed endpoints. The table highlights that similar metrics do not behave identically across documentation contexts.}
\Description{Table comparing endpoint percentage changes for focal metrics in HIL and GitHub corpora. It indicates whether each metric moves in the same direction across corpora and summarizes the interpretation.}
\label{tab:cross_corpus_signal_comparison}
\centering
\scriptsize
\setlength{\tabcolsep}{3pt}
\resizebox{\columnwidth}{!}{%
\begin{tabular}{lcccp{0.30\columnwidth}}
\toprule
\textbf{Metric} & \textbf{HIL $\Delta$\%} & \textbf{GH $\Delta$\%} & \textbf{Dir.?} & \textbf{Interpretation} \\
\midrule
Entity density & -3.5\% & -0.5\% & partial & stable in both; stronger movement in HIL \\
Imperative verb ratio & -23.2\% & -13.2\% & yes & declines in both, but more sharply in HIL \\
Vagueness density & +9.7\% & -3.5\% & no & increases in HIL, decreases in GitHub \\
Constraint density & +18.2\% & -1.0\% & no & rises in HIL, nearly flat in GitHub \\
\bottomrule
\end{tabular}
}
\end{table}

Table~\ref{tab:cross_corpus_signal_comparison} complements the correlation results by showing how the same focal metrics move differently across the two corpora studied. This makes the corpus-specific pattern easier to read than prose alone: the HIL corpus shows stronger procedural and pragmatic movement, whereas GitHub remains comparatively stable or shifts in different directions.

\subsection{Temporal stability of key signals}

\begin{table}[t]
\caption{Temporal stability summary for the HIL corpus using coefficient of variation across weekly aggregate means. Lower CV indicates greater stability relative to the other signals in this study; we do not treat these values as universal cutoffs.}
\Description{Table summarizing HIL metric stability using Week 1 value, Week 17 value, net percentage change, coefficient of variation, and interpretation. Entity density is most stable and imperative verb ratio has the highest variability.}
\label{tab:temporal_stability}
\centering
\footnotesize
\setlength{\tabcolsep}{4pt}
\begin{tabular}{lccccl}
\toprule
\textbf{Metric} & \textbf{Week 1} & \textbf{Week 17} & \textbf{Net \%} & \textbf{CV} & \textbf{Interpretation} \\
\midrule
Entity density & 0.063 & 0.061 & -3.5\% & 0.013 & most stable \\
Vagueness density & 8.11 & 8.89 & +9.7\% & 0.033 & low variability \\
Constraint density & 6.13 & 7.25 & +18.2\% & 0.061 & moderate variability \\
Imperative verb ratio & 0.084 & 0.065 & -23.2\% & 0.080 & highest variability \\
\bottomrule
\end{tabular}
\end{table}

Temporal stability estimates for the HIL corpus further distinguish signals with lower and higher variation. We use CV as a comparative within-study measure over weekly aggregate means rather than as an absolute threshold for practical significance. Directional change and temporal variability capture distinct aspects of documentation revision and should be interpreted separately. Under that interpretation, entity density showed the lowest coefficient of variation (CV $\approx 0.013$), indicating the greatest stability across weekly snapshots. Vagueness density increased directionally but remained relatively smooth over time (CV $\approx 0.033$), while constraint density showed moderate variability (CV $\approx 0.061$). Imperative ratio showed the highest variability (CV $\approx 0.080$) among the four signals. These results indicate that revision sensitivity is not captured by endpoint movement alone: a signal may shift directionally while remaining relatively smooth across weekly snapshots. Distinguishing endpoint movement (net \%) from temporal variability (CV) is necessary to avoid conflating directional change with volatility: the former indicates direction of change, whereas the latter summarizes variability over time.

\subsection{Illustrative Application to Public Documentation}
\label{sec:results:github}

To illustrate how the same measurements behave outside the operational setting, the metrics were applied to a public corpus: the official GitHub documentation repository\cite{githubDocs}\footnote{GitHub Docs is licensed under the Creative Commons Attribution 4.0 International License (CC BY 4.0).}. Annual snapshots from January 13th in 2024, 2025, and 2026 were analyzed; exact commit hashes are provided in Appendix~\ref{app:github}.

The GitHub corpus grew from 5,870 documents in 2024 to 7,175 in 2026 (+22.2\%). Mean words per document increased modestly (+5.3\%), mean sentence count increased (+15.8\%), and average sentence length decreased slightly ($-3.2$\%).

Linguistic and procedural metrics exhibited patterns distinct from the operational runbook corpus. Entity density and imperative verb ratio remained stable, vagueness density decreased modestly, and constraint density showed minimal change. Structural indicators differed more substantially, with list items per document increasing (+26.5\%), reflecting greater use of explicitly structured list content. Observed differences across the two corpora suggest that metric behavior is sensitive to repository context rather than governed by uniform thresholds.

Year-over-year trends showed year-specific variations without sustained directional change. Full metric tables and yearly trend breakdowns are provided in Appendix~\ref{app:github}.

\section{Discussion}

The measurement approach presented here supports monitoring of version-controlled documentation repositories through three complementary signal roles. Stable anchors provide a baseline for comparing snapshots without overreacting to routine edits. Revision-sensitive indicators register localized reformulation and help identify where maintenance activity is concentrating. Context-dependent signals require interpretation relative to corpus context and revision regime rather than universal thresholds. In that sense, the contribution is diagnostic rather than evaluative: the metrics support revision monitoring and governance, but they do not by themselves establish whether a revision is globally ``better'' or whether it improves downstream task success.

Structural indicators remained largely stable across the 17-week study period. Document length, sentence length, and hierarchical organization showed limited variation, indicating that updates did not occur primarily through broad expansion or restructuring. This makes them useful baseline signals for repository-level monitoring and helps isolate changes expressed through localized additions and reformulations.

In contrast, several linguistic metrics exhibited directional change. The imperative verb ratio declined, and constraint and obligation language increased. These shifts are consistent with changes in how procedural actions are framed rather than changes in procedural scope.

From a measurement perspective, these trends indicate that revisions affected multiple aspects of procedural language simultaneously. The correlation structure is informative here: in the HIL corpus, imperative and constraint language were weakly positively associated rather than opposed, and constraint and vagueness markers showed moderate co-variation. These relationships suggest overlapping patterns of revision rather than a tradeoff in which one metric systematically replaces another.

These distinctions constrain interpretation because linguistic form influences how instructions are read and acted upon. Imperative constructions support direct execution, while conditional and obligation-based language introduces contextual flexibility at the cost of immediacy \citep{de-marneffe-etal-2012-happen}. The observed trends therefore suggest changes in how procedures were expressed rather than simple increases or decreases in directive content. 

Named entity density remained stable across the study period, constraining interpretation of the observed linguistic shifts. Stability in referential grounding indicates that changes in directive and pragmatic framing were not driven by a broad move toward abstraction or generic phrasing. While this does not rule out localized instances of vague content, it establishes a corpus-level baseline against which longer-term deviations can be meaningfully evaluated.

Markers of vagueness and hedging also increased over time, alongside rising constraint language. In procedural documentation, such markers may reflect context-sensitive conditional guidance or introduce ambiguity depending on usage. The corrected temporal summaries suggest that vagueness should not be treated as the most volatile signal here: its increase was comparatively smooth, constraint language showed moderate variability, and imperative framing varied most across weekly snapshots. These signals are therefore best treated as context-dependent rather than uniformly positive or negative. Their analytical value lies in flagging regions of potential ambiguity for closer inspection rather than in characterizing revision quality.

Not all signals exhibited the same level of interpretive reliability under longitudinal analysis. Metrics such as list density and code fence counts fluctuated in response to localized edits and document additions, limiting their usefulness for tracking sustained revision dynamics. In contrast, directive and obligation-related signals exhibited coherent trajectories over time, making them better suited for monitoring change without conflating localized formatting updates with substantive reformulation.

Comparison with a public technical documentation corpus further illustrates differences across the two corpora studied. Differences across corpora are more pronounced in the structure of relationships among signals than in their individual magnitudes. GitHub product documentation exhibited growth in explicit structural elements, while the operational runbooks analyzed here showed stronger shifts in directive and pragmatic framing. Some relationships, such as the negative association between vagueness and entity density, recur across corpora, but others weaken substantially or disappear outside the HIL setting. The comparison should not be read as a normative benchmark in which one corpus defines desired values for the other.

Overall, repository change in this setting is more clearly reflected in linguistic reformulation than in volume or structure. Structural and referential signals function as stable anchors, with entity density in particular serving as a stable anchor across weekly snapshots. Imperative framing is the most variable of the focal linguistic signals, while vagueness increases smoothly and constraint language occupies a middle position that requires contextual interpretation.

\subsection{Boundaries and possible downstream implications}

This study does not evaluate downstream use directly, but the observed metric patterns motivate bounded hypotheses for later work on retrieval and documentation-supported guidance. Declining imperative density may alter how directly action sequences are expressed. Increases in vagueness markers may change how often passages require contextual interpretation. Stable entity density suggests that referential grounding remains comparatively intact even as directive framing changes. 

Within the scope of the present paper, these implications should be treated as hypotheses rather than validated claims. Their value in the present paper is to connect the measurement results to plausible uses without collapsing monitoring into outcome evaluation. Establishing which signals behave reliably under revision is a prerequisite for later studies that test whether those signals correlate with human judgments or task performance.

\section{Conclusion}

This work presented a lightweight measurement approach for monitoring revision in version-controlled operational documentation. By applying transparent linguistic and structural metrics to revision snapshots, the paper characterized patterns of change, association, and stability under ongoing human-authored revision.

The results show that while core volume and structural properties remain stable, repository revision is more clearly reflected in shifts in directive framing and obligation language. Correlation analysis indicates mixed patterns of association, including both positive and negative relationships, with effect sizes that are mostly weak to moderate and not identical across the two corpora studied. Entity density served as a stable anchor for baseline monitoring, imperative usage showed the greatest temporal variability, and vagueness and constraint signals exhibited directional movement while requiring context-sensitive interpretation. Comparison with a public technical documentation corpus further showed that metric behavior differs across the two corpora studied.

Future work can examine how the metrics identified here relate to external criteria such as expert judgment, task-level evaluation, and downstream system use. An illustrative example is provided in Appendix~\ref{app:zero_shot}, showing how such metrics may be summarized by automated systems and highlighting the potential for overinterpretation. More broadly, the framework may support documentation governance, maintenance planning, refinement workflows, and the development of evaluation rubrics for evolving document collections.

\section{Limitations}

This study has several limitations that should be considered when interpreting the results. First, the analysis is observational and confined to a single deployment within one organization. The HIL corpus should therefore be understood as a bounded documentation program observed densely over time, not as a broad population sample. Documentation practices, operational norms, and revision behaviors may differ substantially across organizations, limiting the generalizability of specific trends observed here. The GitHub comparison helps separate recurring from corpus-specific relationships, but it does not eliminate the study's bounded organizational scope.

Second, the longitudinal dataset is densely observed rather than statistically independent at the row level. The same documents appear repeatedly across weekly snapshots, so repeated observations reduce the effective sample size for row-level analyses and require conservative interpretation of correlations and related significance tests.

Third, the study relies on general-purpose natural language processing tools. Named entity recognition and part-of-speech tagging were performed using a spaCy model trained on mixed-domain English text rather than operational documentation. This was an intentional design choice in favor of transparent, reproducible signals that can be computed consistently across corpora. While the model is not optimized for runbooks, the analysis emphasizes relative change and aggregate behavior rather than absolute tagging accuracy. Systematic tagging biases would therefore be expected to affect snapshots consistently, reducing their impact on longitudinal trends, but domain-specific models may yield different absolute values.

Fourth, the NLP signals used here are intentionally coarse. Entity counts, imperative ratios, pronoun rates, and lexicon-based vagueness or constraint markers provide transparent and reproducible proxies, but they do not capture deeper semantic properties such as factual correctness, procedural completeness, or whether a conditional statement is appropriately calibrated to system variability. In operational text, some apparent ``noise'' may also be functional: abbreviated commands, product-specific identifiers, fragmented syntax, and telegraphic phrasing are common in runbooks and may reduce the apparent accuracy of general-purpose tokenization, tagging, and sentence segmentation without undermining the practical usefulness of the documentation for human operators.

Fifth, the lexicons used to identify vagueness and hedging markers are derived from prior work in general English and narrative or advisory text, and have not been formally validated for technical documentation or operational runbooks. Accordingly, these metrics are treated as descriptive signals rather than direct measures of underspecification or quality. Their interpretation is necessarily contextual and should be considered alongside complementary indicators such as constraint language and referential grounding.

Sixth, the dataset spans a fixed and relatively short time horizon corresponding to a single deployment period. Documentation changes are driven by operational events, and longer-term or cyclical patterns may not be captured. In addition, the study does not employ controlled experimental manipulation or counterfactual baselines, precluding causal attribution between the deployed system and observed documentation changes.

Finally, the analysis does not incorporate human judgments of documentation quality or operational effectiveness. While the selected metrics are grounded in prior work and chosen for interpretability, they do not directly assess semantic correctness, completeness, or practical utility. Nor do we validate the automated signals against expert annotation in the present study. For that reason, the present metrics are best understood as diagnostic measurements that can guide inspection and future validation rather than as standalone evaluative criteria.

\section{Ethical Considerations}

This study analyzes documentation maintained under human oversight. All documentation edits were reviewed and approved through established workflows before commit, and responsibility for operational decisions remained with on-call engineers. Potential risks include over-reliance on automated metrics without human judgment and documentation degradation affecting incident response time. Mitigations include: (1) metrics are descriptive, not evaluative; (2) all edits underwent standard review; and (3) feedback mechanisms exist for engineers to flag unhelpful documentation. The analysis does not validate downstream operational performance or decision quality.

\begin{acks}
Large language models were used as writing assistance tools during manuscript preparation, including support for drafting, editing, and structural refinement of prose. All methodological design, data processing, statistical analysis, and interpretation were performed and verified by the authors. No generative model was used to produce experimental results or empirical findings.
\end{acks}

\section*{Data Availability}

The operational documentation corpus analyzed in this study is derived from proprietary materials and cannot be publicly released. A public reference corpus based on the GitHub Docs repository is used for illustrative application of the methodology; exact commit hashes and snapshot details are provided in Appendix~\ref{app:github}. All metric definitions and computation procedures are described in the main text and Appendix~\ref{app:metrics} to support independent reimplementation.

\bibliographystyle{ACM-Reference-Format}
\bibliography{references}

\clearpage
\appendix
\setlength{\floatsep}{2pt plus 1pt minus 1pt}
\setlength{\textfloatsep}{2pt plus 1pt minus 1pt}
\setlength{\intextsep}{2pt plus 1pt minus 1pt}
\setlength{\abovecaptionskip}{1pt}
\setlength{\belowcaptionskip}{4pt}
\renewcommand{\arraystretch}{0.86}

\section{Metric Tables}
\label{app:metric_tables}

This appendix provides supporting metric tables referenced in the main text.
The main paper reports inline summaries of key patterns; the tables below provide detailed values for verification.

\FloatBarrier
\subsection{HIL Corpus: Endpoint and Weekly Summary Tables}
\label{app:descriptive}

\IfFileExists{tables/tab_corpus_volume_w1_w17_part1.tex}{\begin{table}[H]
\caption{HIL corpus volume (Part 1): Document counts and word statistics at Week 1 and Week 17.}
\Description{Table comparing HIL document counts, total words, and mean and median words per document at Week 1 and Week 17.}
\label{tab:corpus_volume_w1_w17_part1}
\footnotesize
\begin{tabular}{lrrrr}
\toprule
Label & Valid & Total & W/Doc & W/Doc \\
      & Docs  & Words & (Mean) & (Med) \\
\midrule
Week 1  & 57 & 51,920 & 910.88 & 418.00 \\
Week 17 & 78 & 70,703 & 906.45 & 450.00 \\
\bottomrule
\end{tabular}
\end{table}
}{\IfFileExists{tables/tab_corpus_volume_w1_w17_part1}{}{}}
\IfFileExists{tables/tab_corpus_volume_w1_w17_part2.tex}{\begin{table}[H]
\caption{HIL corpus volume (Part 2): Sentence statistics and document count change from Week 1 to Week 17.}
\Description{Table comparing HIL mean sentences per document, average sentence length, and document count change at Week 1 and Week 17.}
\label{tab:corpus_volume_w1_w17_part2}
\footnotesize
\begin{tabular}{lrrr}
\toprule
Label & Sent/Doc & Avg Sent & Docs \\
      & (Mean) & Length & Change \\
\midrule
Week 1  & 52.26 & 26.37 & -- \\
Week 17 & 50.74 & 26.61 & +36.8\% \\
\bottomrule
\end{tabular}
\end{table}
}{\IfFileExists{tables/tab_corpus_volume_w1_w17_part2}{}{}}

\IfFileExists{tables/tab_linguistic_procedural_w1_w17_part1.tex}{
\begin{table}[H]
\caption{HIL linguistic metrics (Part 1): Entity Density, Imperative Ratio, Vagueness Density, Constraint Density at Week 1 and Week 17.}
\Description{Table comparing HIL entity density, imperative ratio, vagueness density, and constraint density at Week 1 and Week 17.}
\label{tab:linguistic_w1_w17_part1}
\footnotesize
\begin{tabular}{lrrrr}
\toprule
Label & Entity & Imperative & Vagueness & Constraint \\
      & Density & Ratio & Density & Density \\
\midrule
Week 1  & 0.06 & 0.08 & 8.11 & 6.13 \\
Week 17 & 0.06 & 0.06 & 8.89 & 7.25 \\
\bottomrule
\end{tabular}
\end{table}
}{\IfFileExists{tables/tab_linguistic_procedural_w1_w17_part1}{
\begin{table}[H]
\caption{HIL linguistic metrics (Part 1): Entity Density, Imperative Ratio, Vagueness Density, Constraint Density at Week 1 and Week 17.}
\Description{Table comparing HIL entity density, imperative ratio, vagueness density, and constraint density at Week 1 and Week 17.}
\label{tab:linguistic_w1_w17_part1}
\footnotesize
\begin{tabular}{lrrrr}
\toprule
Label & Entity & Imperative & Vagueness & Constraint \\
      & Density & Ratio & Density & Density \\
\midrule
Week 1  & 0.06 & 0.08 & 8.11 & 6.13 \\
Week 17 & 0.06 & 0.06 & 8.89 & 7.25 \\
\bottomrule
\end{tabular}
\end{table}
}{}}
\IfFileExists{tables/tab_linguistic_procedural_w1_w17_part2.tex}{
\begin{table}[H]
\caption{HIL procedural metrics (Part 2): Code Fences, List Items, Type-Token Ratio, Pronoun Density at Week 1 and Week 17.}
\Description{Table comparing HIL code fence count, list item count, type-token ratio, and pronoun density at Week 1 and Week 17.}
\label{tab:linguistic_w1_w17_part2}
\footnotesize
\begin{tabular}{lrrrr}
\toprule
Label & Code & List & TTR & Pronoun \\
      & Fences & Items & & Density \\
\midrule
Week 1  & 1.19 & 29.39 & 0.40 & 25.18 \\
Week 17 & 1.23 & 28.85 & 0.39 & 25.36 \\
\bottomrule
\end{tabular}
\end{table}
}{\IfFileExists{tables/tab_linguistic_procedural_w1_w17_part2}{
\begin{table}[H]
\caption{HIL procedural metrics (Part 2): Code Fences, List Items, Type-Token Ratio, Pronoun Density at Week 1 and Week 17.}
\Description{Table comparing HIL code fence count, list item count, type-token ratio, and pronoun density at Week 1 and Week 17.}
\label{tab:linguistic_w1_w17_part2}
\footnotesize
\begin{tabular}{lrrrr}
\toprule
Label & Code & List & TTR & Pronoun \\
      & Fences & Items & & Density \\
\midrule
Week 1  & 1.19 & 29.39 & 0.40 & 25.18 \\
Week 17 & 1.23 & 28.85 & 0.39 & 25.36 \\
\bottomrule
\end{tabular}
\end{table}
}{}}

\FloatBarrier
\subsection{HIL Corpus: Week-by-Week Trends}
\label{app:hil_trends}

\IfFileExists{tables/tab_weekly_trends_part1.tex}{
\begin{table}[H]
\caption{Weekly trends (Part 1): Linguistic metrics. Percentage change from baseline (Week 1) for Entity Density, Imperative Ratio, Vagueness Density, Constraint Density.}
\Description{Table listing percentage change from Week 1 for HIL linguistic metrics at Weeks 1, 5, 9, 13, and 17.}
\label{tab:weekly_trends_part1}
\footnotesize
\begin{tabular}{lrrrr}
\toprule
Week & Entity & Imperative & Vagueness & Constraint \\
     & Density & Ratio & Density & Density \\
\midrule
Week 1  & +0.0\% & +0.0\%  & +0.0\% & +0.0\% \\
Week 5  & -2.3\% & -15.7\% & +6.8\% & +18.9\% \\
Week 9  & -2.3\% & -18.8\% & +6.6\% & +22.5\% \\
Week 13 & -3.4\% & -23.2\% & +9.4\% & +16.9\% \\
Week 17 & -3.5\% & -23.2\% & +9.7\% & +18.2\% \\
\bottomrule
\end{tabular}
\end{table}
}{\IfFileExists{tables/tab_weekly_trends_part1}{}{}}
\IfFileExists{tables/tab_weekly_trends_part2.tex}{
\begin{table}[H]
\caption{Weekly trends (Part 2): Structural metrics. Percentage change from baseline (Week 1) for Code Fences, List Items, Average Sentence Length, Document Count.}
\Description{Table listing percentage change from Week 1 for HIL structural metrics and document count at Weeks 1, 5, 9, 13, and 17.}
\label{tab:weekly_trends_part2}
\footnotesize
\begin{tabular}{lrrrr}
\toprule
Week & Code & List & Avg Sent & Docs \\
     & Fences & Items & Length & \\
\midrule
Week 1  & +0.0\% & +0.0\% & +0.0\% & +0.0\% \\
Week 5  & -11.2\% & -5.2\% & -2.0\% & +19.3\% \\
Week 9  & +7.9\% & +0.1\% & +1.9\% & +28.1\% \\
Week 13 & +1.0\% & -1.5\% & +1.9\% & +36.8\% \\
Week 17 & +3.2\% & -1.8\% & +0.9\% & +36.8\% \\
\bottomrule
\end{tabular}
\end{table}
}{\IfFileExists{tables/tab_weekly_trends_part2}{}{}}


\FloatBarrier
\subsection{Public Verification Corpus: GitHub Documentation}
\label{app:github}

\IfFileExists{tables/tab_github_corpus_volume_part1.tex}{\begin{table}[H]
\caption{GitHub corpus volume (Part 1): Document counts and word statistics for 2024 and 2026.}
\Description{Table comparing GitHub documentation valid document counts, total words, and mean and median words per document for 2024 and 2026.}
\label{tab:github_corpus_volume_part1}
\footnotesize
\begin{tabular}{lrrrr}
\toprule
Year & Valid & Total & W/Doc & W/Doc \\
     & Docs  & Words & (Mean) & (Med) \\
\midrule
2024 & 5,870 & 1,663,685 & 283.42 & 68.00 \\
2026 & 7,175 & 2,141,098 & 298.41 & 72.00 \\
\bottomrule
\end{tabular}
\end{table}
}{\IfFileExists{tables/tab_github_corpus_volume_part1}{}{}}
\IfFileExists{tables/tab_github_corpus_volume_part2.tex}{\begin{table}[H]
\caption{GitHub corpus volume (Part 2): Sentence statistics and document count change from 2024 to 2026.}
\Description{Table comparing GitHub documentation mean sentences per document, average sentence length, and document count change for 2024 and 2026.}
\label{tab:github_corpus_volume_part2}
\footnotesize
\begin{tabular}{lrrr}
\toprule
Year & Sent/Doc & Avg Sent & Docs \\
     & (Mean) & Length & Change \\
\midrule
2024 & 15.81 & 34.93 & -- \\
2026 & 18.30 & 33.80 & +22.2\% \\
\bottomrule
\end{tabular}
\end{table}
}{\IfFileExists{tables/tab_github_corpus_volume_part2}{}{}}

\IfFileExists{tables/tab_github_linguistic_procedural_part1.tex}{
\begin{table}[H]
\caption{GitHub linguistic metrics (Part 1): Entity Density, Imperative Ratio, Vagueness Density for 2024 and 2026.}
\Description{Table comparing GitHub documentation entity density, imperative ratio, and vagueness density for 2024 and 2026.}
\label{tab:github_linguistic_part1}
\footnotesize
\begin{tabular}{lrrr}
\toprule
Year & Entity & Imperative & Vagueness \\
     & Density & Ratio & Density \\
\midrule
2024 & 0.02 & 0.03 & 17.16 \\
2026 & 0.02 & 0.03 & 16.57 \\
\bottomrule
\end{tabular}
\end{table}
}{\IfFileExists{tables/tab_github_linguistic_procedural_part1}{
\begin{table}[H]
\caption{GitHub linguistic metrics (Part 1): Entity Density, Imperative Ratio, Vagueness Density for 2024 and 2026.}
\Description{Table comparing GitHub documentation entity density, imperative ratio, and vagueness density for 2024 and 2026.}
\label{tab:github_linguistic_part1}
\footnotesize
\begin{tabular}{lrrr}
\toprule
Year & Entity & Imperative & Vagueness \\
     & Density & Ratio & Density \\
\midrule
2024 & 0.02 & 0.03 & 17.16 \\
2026 & 0.02 & 0.03 & 16.57 \\
\bottomrule
\end{tabular}
\end{table}
}{}}
\IfFileExists{tables/tab_github_linguistic_procedural_part2.tex}{
\begin{table}[H]
\caption{GitHub procedural metrics (Part 2): Constraint Density, Code Fences, List Items for 2024 and 2026.}
\Description{Table comparing GitHub documentation constraint density, code fence count, and list item count for 2024 and 2026.}
\label{tab:github_linguistic_part2}
\footnotesize
\begin{tabular}{lrrr}
\toprule
Year & Constraint & Code & List \\
     & Density & Fences & Items \\
\midrule
2024 & 9.55 & 0.63 & 4.77 \\
2026 & 9.45 & 0.65 & 6.04 \\
\bottomrule
\end{tabular}
\end{table}
}{\IfFileExists{tables/tab_github_linguistic_procedural_part2}{
\begin{table}[H]
\caption{GitHub procedural metrics (Part 2): Constraint Density, Code Fences, List Items for 2024 and 2026.}
\Description{Table comparing GitHub documentation constraint density, code fence count, and list item count for 2024 and 2026.}
\label{tab:github_linguistic_part2}
\footnotesize
\begin{tabular}{lrrr}
\toprule
Year & Constraint & Code & List \\
     & Density & Fences & Items \\
\midrule
2024 & 9.55 & 0.63 & 4.77 \\
2026 & 9.45 & 0.65 & 6.04 \\
\bottomrule
\end{tabular}
\end{table}
}{}}

\IfFileExists{tables/tab_github_yearly_trends_part1.tex}{
\begin{table}[H]
\caption{GitHub yearly trends (Part 1): Linguistic metrics. Percentage change from 2024 baseline for Entity Density, Imperative Ratio, Vagueness Density, Constraint Density.}
\Description{Table listing percentage change from 2024 baseline for GitHub linguistic metrics in 2024, 2025, and 2026.}
\label{tab:github_yearly_trends_part1}
\footnotesize
\begin{tabular}{lrrrr}
\toprule
Year & Entity & Imperative & Vagueness & Constraint \\
     & Density & Ratio & Density & Density \\
\midrule
2024 & +0.0\% & +0.0\%  & +0.0\% & +0.0\% \\
2025 & -0.3\% & -18.1\% & +1.8\% & +2.6\% \\
2026 & -0.5\% & -13.2\% & -3.5\% & -1.0\% \\
\bottomrule
\end{tabular}
\end{table}
}{\IfFileExists{tables/tab_github_yearly_trends_part1}{}{}}
\IfFileExists{tables/tab_github_yearly_trends_part2.tex}{
\begin{table}[H]
\caption{GitHub yearly trends (Part 2): Structural metrics. Percentage change from 2024 baseline for Code Fences, List Items, Average Sentence Length, Document Count.}
\Description{Table listing percentage change from 2024 baseline for GitHub structural metrics and document count in 2024, 2025, and 2026.}
\label{tab:github_yearly_trends_part2}
\footnotesize
\begin{tabular}{lrrrr}
\toprule
Year & Code & List & Avg Sent & Docs \\
     & Fences & Items & Length & \\
\midrule
2024 & +0.0\% & +0.0\% & +0.0\% & +0.0\% \\
2025 & -2.6\% & +6.0\% & -3.4\% & +9.9\% \\
2026 & +1.9\% & +26.5\% & -3.2\% & +22.2\% \\
\bottomrule
\end{tabular}
\end{table}
}{\IfFileExists{tables/tab_github_yearly_trends_part2}{}{}}

\FloatBarrier
\IfFileExists{tables/tab_github_snapshots.tex}{\begin{table}[t]
\caption{GitHub documentation repository snapshots used for analysis. All snapshots were taken on January 13th of their respective years from the \texttt{github/docs} repository.}
\Description{Table listing the GitHub documentation repository commit hashes and snapshot dates used for the 2024, 2025, and 2026 analyses.}
\label{tab:github_snapshots}
\centering
\small
\begin{tabular}{lll}
\toprule
Year & Commit Hash & Snapshot Date \\
\midrule
2024 & \texttt{62bf6d6} & January 13, 2024 \\
2025 & \texttt{83b29a0} & January 13, 2025 \\
2026 & \texttt{d62681e} & January 13, 2026 \\
\bottomrule
\end{tabular}
\end{table}
}{\IfFileExists{tables/tab_github_snapshots}{\begin{table}[t]
\caption{GitHub documentation repository snapshots used for analysis. All snapshots were taken on January 13th of their respective years from the \texttt{github/docs} repository.}
\Description{Table listing the GitHub documentation repository commit hashes and snapshot dates used for the 2024, 2025, and 2026 analyses.}
\label{tab:github_snapshots}
\centering
\small
\begin{tabular}{lll}
\toprule
Year & Commit Hash & Snapshot Date \\
\midrule
2024 & \texttt{62bf6d6} & January 13, 2024 \\
2025 & \texttt{83b29a0} & January 13, 2025 \\
2026 & \texttt{d62681e} & January 13, 2026 \\
\bottomrule
\end{tabular}
\end{table}
}{}}

\section{Complete Metric Definitions}
\label{app:metrics}

This appendix provides a comprehensive reference of all 39 metrics computed by the RunbookTextStats tool. Metrics are organized into eight categories reflecting distinct properties of technical documentation.

\subsection{File Metrics (3)}

\begin{table}[H]
\Description{Table defining file metrics: bytes, character count, and line count.}
\footnotesize
\begin{tabular}{p{0.4\columnwidth} p{0.45\columnwidth}}
\toprule
\textbf{Metric} & \textbf{Description} \\
\midrule
\texttt{bytes} & File size (bytes) \\
\texttt{chars} & Character count \\
\texttt{lines} & Line count \\
\bottomrule
\end{tabular}
\end{table}

\subsection*{Markdown Structure (13)}

\begin{table}[H]
\Description{Table defining Markdown structure metrics, including header counts, code blocks, list items, and checklist count.}
\footnotesize
\begin{tabular}{p{0.4\columnwidth} p{0.45\columnwidth}}
\toprule
\textbf{Metric} & \textbf{Description} \\
\midrule
\texttt{header\_count} & Total headers (H1--H6) \\
\texttt{header\_h*\_count} & Counts for H1--H6 (6 metrics) \\
\texttt{code\_fence\_count} & Fenced code blocks \\
\texttt{code\_block\_lines} & Lines in code blocks \\
\texttt{bullet\_list\_count} & Bullet list items \\
\texttt{numbered\_list\_count} & Numbered list items \\
\texttt{checklist\_count} & Checklist items \\
\texttt{list\_item\_count} & Total list items \\
\bottomrule
\end{tabular}
\end{table}

\subsection*{Text Volume (3)}

\begin{table}[H]
\Description{Table defining text volume metrics: word count, paragraph count, and regex-based sentence count.}
\footnotesize
\begin{tabular}{p{0.4\columnwidth} p{0.45\columnwidth}}
\toprule
\textbf{Metric} & \textbf{Description} \\
\midrule
\texttt{word\_count} & Word count (whitespace-separated) \\
\texttt{paragraph\_count} & Paragraph count \\
\texttt{sentence\_count\_basic} & Sentence count (regex heuristic) \\
\bottomrule
\end{tabular}
\end{table}

\subsection*{Vagueness \& Constraint}

\begin{table}[H]
\Description{Table defining vagueness and constraint density metrics and listing the lexical marker sets used for each.}
\footnotesize
\setlength{\tabcolsep}{6pt}
\begin{tabular}{p{0.38\columnwidth} p{0.52\columnwidth}}
\toprule
\textbf{Metric} & \textbf{Definition and Marker Set} \\
\midrule
\texttt{vagueness\_density} &
Lexical markers of uncertainty and imprecision, computed as occurrences per 1{,}000 words.
The 30 markers include epistemic modals and hedge expressions:
\emph{may, might, could, would, should, can, possibly, perhaps, probably, likely, somewhat, relatively, fairly, rather, quite, sort of, kind of, apparently, seemingly, arguably, roughly, approximately, about, around, generally, typically, often, sometimes, usually, mostly}. \\
\addlinespace
\texttt{constraint\_density} &
Lexical markers of obligation, prohibition, and conditional constraint, computed as occurrences per 1{,}000 words.
The 28 markers include directive modals, prohibitive terms, and fixed expressions:
\emph{must, shall, will, need, needs, needed, require, requires, required, necessary, never, avoid, prohibited, forbidden, do not, don't, must not, mustn't, shall not, shan't, cannot, can't, need to, have to, has to, had to, required to, necessary to, only if, unless, provided that, if and only if}. \\
\bottomrule
\end{tabular}
\end{table}

\subsection*{Basic NLP Metrics (5)}

\emph{Blank or trained spaCy models}

\begin{table}[H]
\Description{Table defining basic NLP metrics, including token count, sentence count, average sentence length, type-token ratio, and moving average type-token ratio.}
\footnotesize
\begin{tabular}{p{0.5\columnwidth} p{0.38\columnwidth}}
\toprule
\textbf{Metric} & \textbf{Description} \\
\midrule
\texttt{token\_count} & Token count (excl. spaces) \\
\texttt{sentence\_count} & Sentence count \\
\texttt{avg\_sentence\_len\_tokens} & Avg tokens per sentence \\
\texttt{type\_token\_ratio} & Lexical diversity (unique/total) \\
\texttt{moving\_avg\_ttr} & Length-independent diversity \\
\bottomrule
\end{tabular}
\end{table}

\subsection*{Named Entity Recognition (6)}

\emph{Requires trained spaCy NER}

\begin{table}[H]
\Description{Table defining named entity recognition metrics, including entity count, entity density, unique entities, entity mention ratio, entity type counts, and top entities.}
\footnotesize
\begin{tabular}{p{0.4\columnwidth} p{0.5\columnwidth}}
\toprule
\textbf{Metric} & \textbf{Description} \\
\midrule
\texttt{entity\_count} & Total named entities \\
\texttt{entity\_density} & Entities per token (0--1) \\
\texttt{unique\_entity\_count} & Unique entity strings \\
\texttt{entity\_mention\_ratio} & Unique/total entities (0--1) \\
\texttt{entity\_type\_counts} & Entity types dict \\
\texttt{top\_entities} & Most frequent entities \\
\bottomrule
\end{tabular}
\end{table}

\subsection*{Part-of-Speech (3)}

\emph{Requires trained spaCy POS tagger}

\begin{table}[H]
\Description{Table defining part-of-speech metrics: pronoun count, pronoun density, and imperative verb ratio.}
\footnotesize
\begin{tabular}{p{0.4\columnwidth} p{0.45\columnwidth}}
\toprule
\textbf{Metric} & \textbf{Description} \\
\midrule
\texttt{pronoun\_count} & Total pronouns \\
\texttt{pronoun\_density} & Pronouns per 1K tokens \\
\texttt{imperative\_verb\_ratio} & Imperative/all verbs (0--1) \\
\bottomrule
\end{tabular}
\end{table}

\subsection*{Metadata (4)}

\begin{table}[H]
\Description{Table defining metadata fields: snapshot identifier, document path hash, relative repository path, and error message.}
\footnotesize
\begin{tabular}{p{0.4\columnwidth} p{0.45\columnwidth}}
\toprule
\textbf{Metric} & \textbf{Description} \\
\midrule
\texttt{snapshot\_id} & Snapshot identifier \\
\texttt{doc\_id} & Document path hash \\
\texttt{repo\_path} & Relative path \\
\texttt{error} & Error message (if failed) \\
\bottomrule
\end{tabular}
\end{table}

\paragraph{Metric Selection Rationale}

These metrics are selected to capture complementary dimensions of technical documentation without requiring task-specific models or manual annotation:

\begin{itemize}[topsep=0pt,itemsep=2pt]
\item \textbf{Structural organization} (Markdown structure, text volume) reflects navigability and hierarchical organization
\item \textbf{Procedural explicitness} (list items, code blocks) indicates step-oriented guidance
\item \textbf{Linguistic properties} (lexical diversity, sentence length) capture writing complexity
\item \textbf{Referential grounding} (entity density, pronoun usage) indicates concrete vs. underspecified content
\item \textbf{Uncertainty and obligation} (vagueness/constraint density) capture actionability in procedural text
\end{itemize}

All metrics are computed automatically from markdown source without semantic interpretation, enabling longitudinal analysis at repository scale.

\section{Illustrative Example of Automated Interpretation (Non-Evaluated)}
\label{app:zero_shot}

This appendix provides an illustrative example of how summary statistics derived from documentation metrics may be interpreted by an automated system. It is included to demonstrate the potential for overinterpretation when such signals are treated as evaluative rather than descriptive.

The example does not constitute a validated evaluation method. The model output is not used as evidence in this study, and no claims are made about its correctness or reliability. In particular, qualitative judgments such as ``improving'' should not be taken as supported conclusions of the analysis.

\begin{quote}
\ttfamily\footnotesize
You are given summary metrics comparing documentation snapshots over time for the 

Task:
Provide an example qualitative summary (non-validated) of the documentation metrics.

Criteria (use only what the metrics support):
- Volume (growth, stability, or contraction)
- Organization / order (structural indicators such as headers, lists, code blocks)
- Specificity (grounding and concreteness proxies such as entity density,
  constraints, vagueness markers)

Instructions:
1) First list the 5–10 metrics you relied on and what each suggests.
2) Then give an example overall summary label.
3) If evidence is mixed, say so and explain why.
4) Do not assume that changes imply quality unless the metrics directly
   support that claim.
5) Identify positive and concerning trends.

Input Metrics: (Derived from Latex Tables Snippets)
Given the following document change statistics:

```
Label, Entity Density, Imperative Ratio, Vagueness Density, Constraint Density
A, 0.06, 0.08, 8.11, 6.13
B, 0.06, 0.06, 8.89, 7.25

Label, Code Fences, List Items, Type-Token Ration, Pronoun Density
A, 1.19, 29.39, 0.40, 25.18
B, 1.23, 28.85, 0.39, 25.36

Label, Valid Docs, Total Words, W/Doc (Mean), W/Doc (Med)
A, 57, 51,920, 910.88, 418.00
B, 78, 70,703, 906.45, 450.00

Label, Sent/Doc (Mean), Avg Sent Length, Docs Change
A, 52.26, 26.37, --
B, 50.74, 26.61, +36.8\%
```
\end{quote}

\subsection{Example Model Response}
Generated with Claude Sonnet 4.5
\begin{quote}
\ttfamily\footnotesize
Example qualitative summary (non-validated)
1. Key Metrics Analysis

Volume Metrics:

    Valid Docs (A: 57 → B: 78, +36.8\%) – Significant growth in documentation coverage, indicating expanded scope or new areas being documented.
    Total Words (A: 51,920 → B: 70,703) – Substantial absolute growth (\textasciitilde36\%) consistent with document count increase.
    W/Doc Mean (A: 910.88 → B: 906.45) – Stable average document length, suggesting growth is through breadth rather than depth inflation.
    W/Doc Median (A: 418 → B: 450) – Slight increase suggests the new documents are reasonably substantive, not just stubs.

Organization Metrics: 5. Code Fences (A: 1.19 → B: 1.23) – Slight increase in code examples per document, maintaining technical illustration. 6. List Items (A: 29.39 → B: 28.85) – Marginal decrease but essentially stable, indicating consistent structural organization.

Specificity Metrics: 7. Constraint Density (A: 6.13 → B: 7.25) – Notable 18\% increase suggests more specific requirements, limits, or rules being documented. 8. Vagueness Density (A: 8.11 → B: 8.89) – Concerning 9.6\% increase in vague language markers. 9. Type-Token Ratio (A: 0.40 → B: 0.39) – Slight decrease indicates marginally more repetitive vocabulary (expected with volume growth). 10. Entity Density (A: 0.06 → B: 0.06) – Stable, showing consistent concrete reference density.
2. Overall Summary: Mixed patterns with caution

This qualitative summary is produced by the model and is not supported or endorsed by the analysis presented in this paper.
3. Mixed Evidence Explanation

Positive trends:

    37\% expansion in coverage without diluting document substance (stable W/Doc metrics)
    Increased constraint density signals more precise, actionable documentation
    Structural elements (code fences, lists) remain consistent despite rapid growth
    Imperative ratio decreased slightly (0.08 → 0.06), potentially indicating more explanatory vs. purely prescriptive content

Concerning trends:

    Vagueness density increased 9.6\%, working against the gains in constraint specificity
    The combination of higher constraints BUT higher vagueness suggests possible inconsistency in documentation quality across the expanded corpus
    Slightly lower type-token ratio may indicate increased boilerplate or templated content

The remainder of this example is omitted here because prescriptive recommendations would overstate what these descriptive metrics can support on their own.
\end{quote}

\end{document}